\documentclass[conference]{IEEEtran}
\IEEEoverridecommandlockouts

\usepackage[hidelinks]{hyperref}
\usepackage{cite}
\usepackage{amsmath,amssymb,amsfonts}
\usepackage{float}
\usepackage{algorithmic}
\usepackage{graphicx}
\usepackage[ruled,noline,linesnumbered]{algorithm2e}
\usepackage{textcomp}
\usepackage{xcolor}
\usepackage{booktabs} 
\usepackage{scalerel} 
\usepackage{caption}
\usepackage{multirow}
\usepackage{comment}
\usepackage{placeins}
\newcommand{\orcidicon}[1]{\href{https://orcid.org/#1}{\includegraphics[width=8pt]{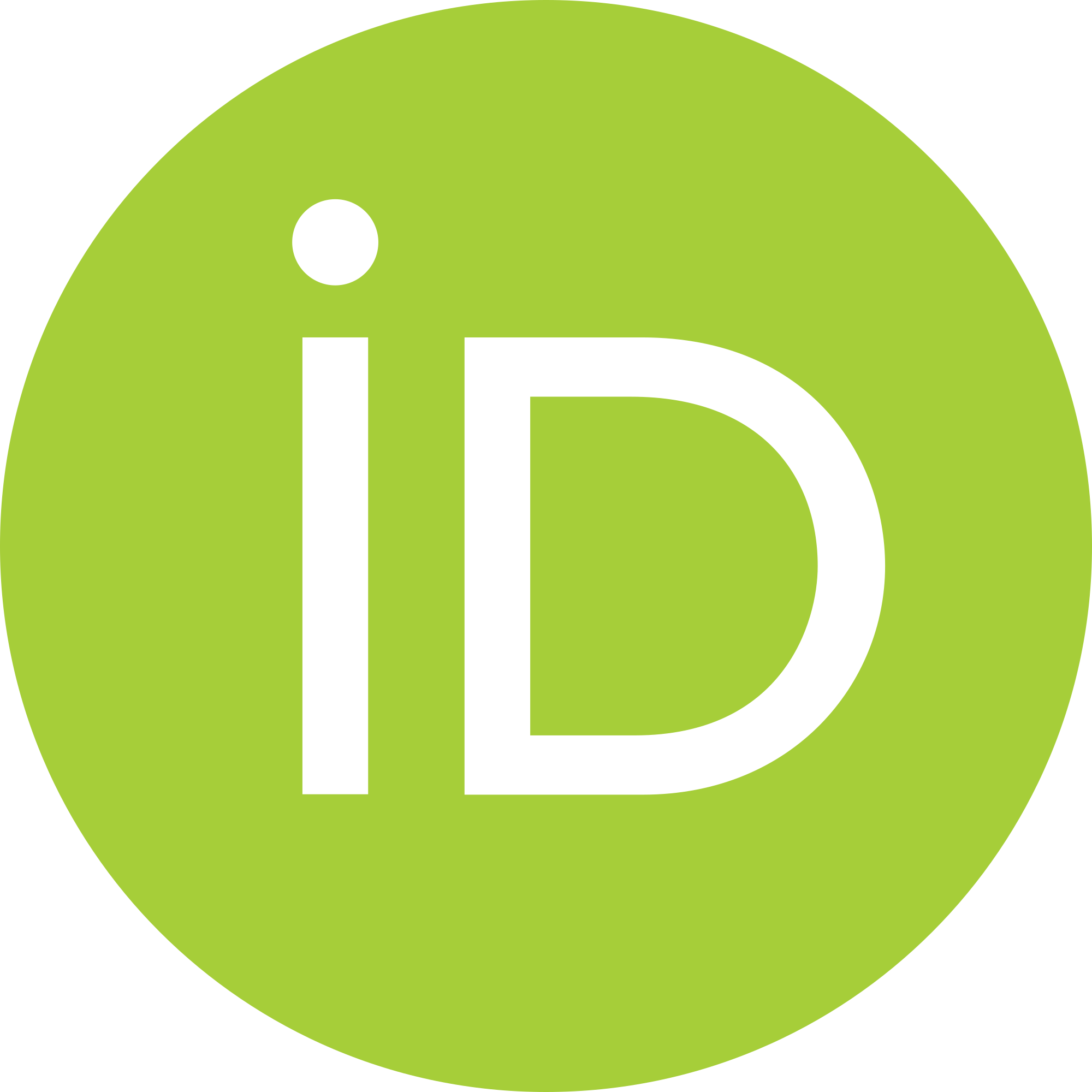}}}
\begin{document}

\title{Hardware Efficient Approximate Convolution with Tunable Error Tolerance for CNNs}

\author{
    \IEEEauthorblockN{
        Vishal Shashidhar\textsuperscript{1} \orcidicon{0009-0002-1309-8134}, 
        Anupam Kumari\textsuperscript{2} \orcidicon{0000-0002-6504-363X}, 
        Roy P. Paily\textsuperscript{3} \orcidicon{0000-0003-3004-9369}
    }
    \IEEEauthorblockA{Department of Electronics and Electrical Engineering, Indian Institute of Technology Guwahati}
    \IEEEauthorblockA{Email: \{v.shashidhar\textsuperscript{1}, a.kumari\textsuperscript{2}, roypaily\textsuperscript{3}\}@iitg.ac.in}
}

\maketitle

\begin{abstract}
Modern Convolutional Neural Networks (CNNs) are increasingly power-intensive and computationally heavy, posing significant challenges for deployment on resource-constrained edge devices. Although current research often exploits "hard" sparsity by skipping mathematical zeros, these methods are inherently limited because the fraction of mathematical zero values decreases significantly in deeper feature maps. Although ReLU activations typically produce a zero fraction of only 20–50\% in these layers, smooth activation functions like tanh generate virtually no zeros at all, making traditional skipping techniques ineffective. This work proposes a "soft sparsity" paradigm that utilizes a novel hardware-efficient approximation strategy to selectively omit multiplications whose contributions to the final output are negligible, even if they are non-zero. The method leverages the Most Significant Bit (MSB) as a low-cost hardware proxy for logarithmic magnitude, allowing the system to compare relative product sizes against a tunable threshold without explicitly performing the multiplication. Integrated as a custom instruction within a 32-bit RISC-V processor, this approach was evaluated using the LeNet-5 architecture on the MNIST dataset. The results demonstrate that for ReLU-based models, inference can be performed by reducing the number of MACs by 88.42\% with no loss of accuracy. Furthermore, the method remains effective for smoother activations such as tanh, which lack hard zeros, achieving a reduction of 74.87\% of total multiplications while maintaining full accuracy. This translates to a 5x reduction in number of multiplication operations relative to traditional hard-zero skipping paradigms. As a consequence of this reduction in multiplication operations that have to be performed, inactive multipliers can be clock gated to conserve power. Power reduction will be sub linear to the reduction in multiplications as prior work has indicated memory access to also be a large contributor to power consumption during convolution. This work estimates power reductions of 35.2\% and 29.96\% per LeNet-5 inference using the ReLU and the tanh activation function, respectively.    
\end{abstract}


\section{Introduction}
CNNs excel at identifying spatial patterns, which has found a wide variety of applications, including object classification in images, facial recognition, NLP tasks, and structured signal analysis. CNNs are composed of convolution layers, activation layers that perform a non-linear function on the neurons, pooling layers responsible for down-sampling, and fully connected layers that flatten 2D feature maps into a 1D vector and connects every input of the layer to every output of the layer. The values of the filter matrices for the convolution layers and the weights of the FC layer are the parameters which are set and stored during training. Convolutional layers account for the vast majority of arithmetic MAC (multiply and accumulate) operations sometimes accounting for over 99\% of FLOPs (floating point operations) such as in ResNet-50 while representing a negligible fraction of the model’s parameters [1] while Fully Connected layers dominate the model’s memory footprint due to dense interconnections [2], however, modern architectures such as ResNet and GoogLeNet mitigate the heavy FC layers, shifting the parameter distribution more toward convolution and global pool [3].  

Modern CNNs are increasingly becoming bulkier and have billions of MAC operations making them power intensive and unsuitable for edge devices, while also containing redundancy in the form of sparsity where there are large fractions of zero or close to zero values in input matrices and weights whose activation result in negligible contribution to the output.
In CNN workloads, input sparsity varies widely by application. Natural images (e.g. ImageNet, CIFAR) are largely dense, with less than 5\% exact zeros, though 20–50\% of pixels may be near-zero in smooth background regions. Handwritten or binary-like inputs (e.g., MNIST, document images) exhibit 60–90\% zero or near-zero pixels. In circuit and EDA applications—such as netlist adjacency matrices, routing congestion maps, and placement grids—input tensors are often 80–99\% sparse, reflecting localized connectivity and activity. Event-based vision, remote sensing, and scientific grids typically show 70–95\% sparsity, often spatially clustered.
\begin{table}[t]
\caption{Parameter Count and Computational Cost of Popular CNN Architectures}
\label{tab:cnn_params_flops}
\centering
\begin{tabular}{l l r r}
\hline
\textbf{Architecture} & \textbf{Layer Type} & \textbf{\# Parameters} & \textbf{\# Operations (FLOPs)} \\
\hline
LeNet-5     & Conv & $\sim$3k   & $\sim$300k \\
            & FC   & $\sim$58k  & $\sim$60k  \\
\hline
AlexNet     & Conv & $\sim$3.7M & $\sim$650M \\
            & FC   & $\sim$58.6M& $\sim$58M  \\
\hline
ResNet-50   & Conv & $\sim$23.5M& $\sim$3.8G \\
            & FC   & $\sim$2M   & $\sim$2M   \\
\hline
GoogLeNet   & Conv & $\sim$5.7M & $\sim$1.5G \\
            & FC   & $\sim$1M   & $\sim$1M   \\
\hline
\end{tabular}
\end{table}

 In literature, this sparsity has been exploited primarily in two ways, the first being pruning the weights which have a negligible contribution to the output and the second by skipping the computation of MAC operations when the input activation is 0. Both these methods do not directly lead to reduction in number of cycles required as in modern hardware MACs are executed in parallel, which means that even though an operation is skipped, that thread will have to wait for the completion of execution of parallel threads before proceeding to the next operation. Specialized hardware relies on storing only the non-zero values in CSR (compressed sparse row) format or CSC (Compressed Sparse Column) format and accessing elements via indexing, but this results in substantial control and power overhead.

However, power consumption can be reduced by clock gating a multiplier when a MAC operation is not necessary. Prior research has shown that the dominant factor in power consumption is data access and not MAC operations, hence energy savings are sub-linear with respect to operation reduction. Han et al. [4] reported 3-5 pJ per 32-bit MAC operation and 5-10 pJ per 32-bit SRAM access for 45nm technology, meaning that reducing MAC operations by 100\% will result in a reduction of power by about 25-50\% if the corresponding memory access is not optimized, assuming data present in SRAM. K. Guo et al. [5] reported that typical energy for 32-bit SRAM access costs twice as much as 32-bit multiplication for 22nm technology, implying 100\% reduction in MAC operations would result in about 33\% power reduction. Chen et al. [6] achieved a 45\% reduction in power by zero skipping with 65nm technology. DRAM access costs multiple orders of magnitude greater than both SRAM access or MAC operation.

Early literature focused on static weight sparsity in FC layers, Han at el. [4] proposed a 3-stage train-prune-retrain pipeline that compresses neural networks by removing weights with magnitudes below a heuristic threshold, followed by iterative fine-tuning to recover lost accuracy. However, these static methods are inherently limited because they cannot adapt to the "dynamic sparsity" that arises from specific input data and require specialized hardware such as Cambricon-X [7] to achieve speedup where the control and indexing cost contribute to a 30-35\% power overhead.

To capture this input-dependent redundancy, researchers developed dynamic accelerators such as Cnvlutin [8] and NullHop [9], which employ runtime zero-skipping to bypass multiplications involving activations nullified by the ReLU function.If the ReLU activation function is not used, subsequent feature maps will contain no zero values. Even when ReLU is used, the fraction of pixels forced to zero may be insignificant—typically only 20–50\% per layer for MNIST images despite an initial input sparsity of 80\%. Furthermore, in the LeNet architecture, this zero fraction tends to decrease even further in deeper layers. These designs are often hampered by the rigid binary nature of "hard" zeros, incurring significant metadata overhead and load imbalance [10] while still being forced to execute any multiplication where both operands are even slightly non-zero. Although advanced dual-side architectures like Sparch [11] and predictive frameworks like SnaPEA [12] attempt to bridge these gaps by identifying ineffectual neurons before they are fully processed, they remain tethered to the assumption that only a mathematical zero can be safely ignored. This research introduces a more flexible "soft sparsity" paradigm, proposing custom hardware that dynamically evaluates the product of a weight and an activation against a tunable threshold without explicitly computing the product. By skipping computations that fall below this significance level, regardless of whether they are exactly zero, this approach allows for a more aggressive exploitation of data redundancy. This is achieved without any indexing overhead and without having to modify the neural network by cycles of pruning and retraining.

This work is organized as follows: Section II explains the background and motivation behind conv\_approx(), the novel approximation operation that this work introduces. Section III explains the algorithmic principle of using the Most Significant Bit (MSB) as a low-cost hardware proxy for logarithmic magnitude to identify and omit insignificant products. Section IV details the hardware implementation, specifically the integration of a custom conv\_approx() instruction into a 32-bit RISC-V processor using a specialized 5-stage Finite State Machine (FSM). Finally, Section V provides a comprehensive error and performance analysis using the LeNet-5 architecture on the MNIST dataset. The primary contributions include a novel hardware-efficient approximation algorithm with a tunable error-tolerance mechanism to balance accuracy and efficiency, and the demonstrated ability to reduce MAC operations by up to 88.42\% for ReLU and 74.87\% for tanh activations during LeNet-5 inference with negligible impact on accuracy.

\section{Background}

The output of each individual convolution between a feature map and a filter matrix can be written as a matrix where each cell is given by a sum of products. For an output location (i,j),
\begin{equation}
Y(i,j) = \sum_{r=0}^{k-1}\sum_{s=0}^{k-1} X(i+r,\,j+s)\,K(r,s)
\end{equation}
 Current literature only skips computation if activation is a mathematical zero; however, there may exist cases where a particular product in a sum of products is insignificant relative to other products. This may be because the activation or weight is relatively small or because there is another product term that dominates over the rest. In such cases, the computation of product terms that have an insignificant contribution to the sum can be skipped. The output will not be the exact mathematical value, but the objective of convolution layers is to identify spatial patterns, hence numerical accuracy can be traded for reduction in computational operations as long as the accuracy of the final output is not compromised. 
 
It can intuitively be seen that there are cases where some products will dominate over the others such as when the filter overlaps against an edge in the feature map, some weights will correspond to high activation values and will likely have large products, whereas some weights will correspond to low activation values and will likely have products which are insignificant relative to the former products; in such a case the multiplication of the weights and activations whose products are insignificant can be skipped. Further elaboration regarding how we decide whether a multiplication can be skipped without explicitly calculating the product is given in a later section.

The decision regarding skipping computation of certain products is taken after considering both activation and weight, as such a decision taken after considering only either one may lead to the decision being incorrect. This method allows us to robustly skip many more computations than just zero-skipping paradigms.
{
\begin{figure}[t]
    \centering
    \includegraphics[width=\columnwidth]{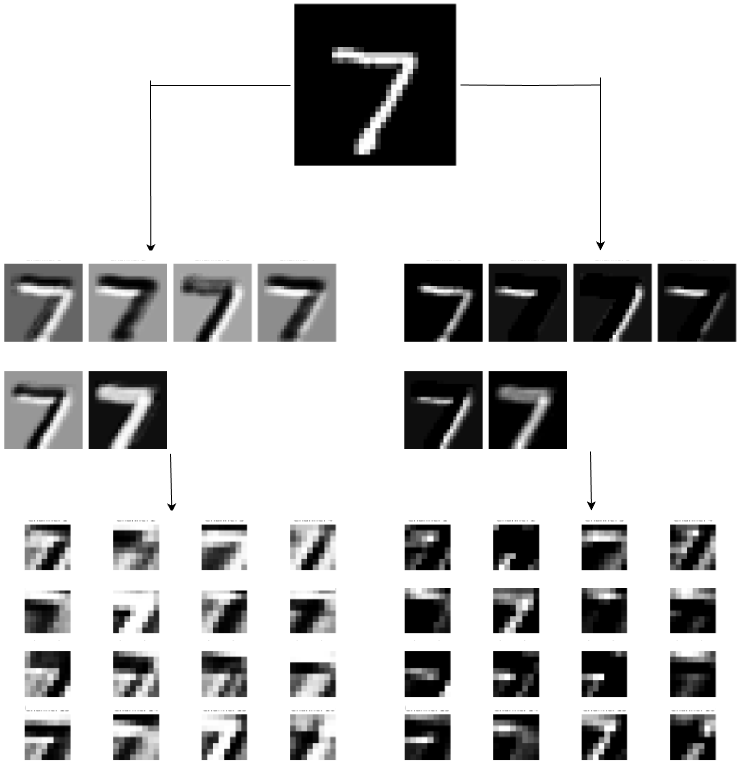}
    \caption{Left and right side show subsequent feature maps with tanh and ReLU activation respectively.}
    \label{fig:example}
\end{figure}
}
Current zero skipping algorithms operate on inputs and subsequent feature maps after application of ReLU activation function, which is basically f(X) = max (0, X). ReLU sets all negative pre-activations to exactly zero, so the neuron stops distinguishing between “slightly negative” and “very negative” inputs in that region. This can remove potentially useful contrast information (e.g. weak vs strong inhibition signals), which might reduce representational capacity compared with smooth or signed activations. Using Techniques like He initialization, batch normalization and residual connections keep activations from collapsing into the negative half line, due to which ReLU-based networks match smoother activations on many vision and language benchmarks. Leaky ReLU, PReLU, and ELU retain a small negative slope, so negative inputs are not completely discarded, and the gradient remains nonzero for all inputs. These variants often show modest but consistent gains. conv\_approx() performs pruning on further feature maps without being conditioned on having to use the ReLU activation function, as it is not dependent on having hard zeros to skip MAC operations.

\section{Algorithmic Principle and Working}
Each element of the output feature map produced by a convolution operation is computed as a weighted sum of input values, i.e., a sum of pairwise products. This work proposes an approximation strategy that reduces the computational cost of convolution by selectively omitting multiplications whose contributions to the final sum are negligible. The core idea is to determine whether the sum of two products, $P_1 + P_2$, with $P_1 \geq P_2$, can be approximated by $P_1$ while ensuring that the resulting error remains below a user-defined threshold. Crucially, this decision is made without explicitly computing either product. By avoiding the multiplication associated with the less significant term, the overall number of MAC operations is reduced. The proposed method relies exclusively on inexpensive hardware operations, enabling efficient implementation. 
{
\includegraphics[width=\columnwidth]{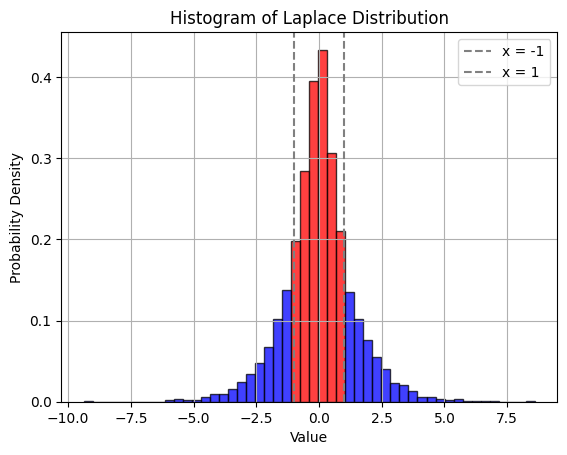}
\captionof{figure}{Histogram of signed magnitude of products during CNN inference. Generated using Laplace distribution.}
\label{fig:example}
}
The above graph shows the data generated by the Laplace distribution with $\mu$ = 0. A Laplace distribution is used because regularization and ReLU activations force the vast majority of weights and activations in a CNN toward zero. When these two zero-centered variables are multiplied, the resulting product distribution becomes even more concentrated at zero, forming a sharp, "spiky" peak. This work proposes detecting the products centered around the origin using MSB location as a hardware proxy for logarithmic value and only performing multiplication if the product is not clustered close to the origin. In other words, approximate the red part of the histogram to be equal to 0 in magnitude. The distance of the vertical threshold lines from the origin can be modified according to the precision required.

The approximation is based on the observation that the position of the most significant bit (MSB) in the binary representation of an integer provides an approximation of its base-2 logarithm; specifically, it is equal to the integer part of $\log_2(x)$. Intuitively, the most significant bit (MSB) indicates the highest power of two that fits within a number, which directly reflects the number’s order of magnitude in base 2. Any positive integer $x$ can be expressed as $x = 2^k + r$, where $2^k$ is the largest power of two not exceeding $x$ and $0 \leq r < 2^k$. The position of the MSB corresponds to this exponent $k$, which means that $\text{MSB}(x) = k$. Taking the base-2 logarithm yields $k \leq \log_2(x) < k+1$, so the position of MSB is exactly the integer part of $\log_2(x)$. Thus, the position of the MSB serves as a simple, hardware-friendly proxy for the logarithmic magnitude of a number.For a product of two values, the MSB position of the result is approximately the sum of the MSB positions of the operands. Consequently, relative magnitudes of products can be compared by examining only sum of MSB positions, without performing multiplication. Let the MSB position of a value $x$ be denoted as $M(x)$. For two products $P_1 = a \cdot b$ and $P_2 = c \cdot d$, define$$\Delta  = (MSB(a) + MSB(b)) - (MSB(c) + MSB(d))$$If $\Delta  \geq T$, for a predefined threshold $T$, then $P_1$ is considered the dominant term, and $P_2$ can be omitted from the computation. MSB extraction and comparison are low-cost hardware operations, making this decision mechanism highly efficient.To illustrate the method, consider approximating $P_1 + P_2$ as $P_1$ under the constraint that $P_2$ contributes no more than 1\% of $P_1$:$$\frac{P_2}{P_1} \leq 0.01$$Taking logarithms yields$$\log_2(P_1) - \log_2(P_2) \geq \log_2(100) \approx 6.64$$Since $MSB(x) \approx \log_2(x)$, this condition can be conservatively approximated using MSB positions as$$(MSB(a) + MSB(b)) - (MSB(c) + MSB(d)) \geq 7$$When this inequality holds, the approximation is applied, requiring only one multiplication instead of two multiplications and one addition.This approximation technique can eliminate a substantial number of multiplications in common convolution scenarios. In convolution neural networks (CNNs), the resulting approximation error has a minimal impact on the accuracy of the inference provided that the extracted features remain sufficiently distinguishable. For example, a convolution with a $3 \times 3$ filter computes an output of the form$$Y = \sum_{i=1}^{9} w_i \cdot x_i$$where $w_i$ are filter coefficients and $x_i$ are input values. In the worst case, the theoretical maximum error introduced by the proposed approximation is eight times the maximum allowed error due to a single omitted product. However, as demonstrated in later sections, empirical results show that the actual error in practical applications is typically much smaller.
\begin{algorithm}[h]
\caption{MSB-Based Approximation for $n$ Products}
\DontPrintSemicolon
\SetKwInOut{Input}{Input}
\SetKwInOut{Output}{Output}

\Input{Sets of operands $\{(a_1, b_1), (a_2, b_2), \dots, (a_n, b_n)\}$, Threshold $T$}
\Output{Approximated sum $S \approx \sum_{i=1}^{n} (a_i \times b_i)$}

\BlankLine
\For{each pair $(a_i, b_i)$}{
    $MSB_i \gets \text{MSB}(a_i) + \text{MSB}(b_i)$\;
}

$MSB_{max} \gets \max(MSB_1, MSB_2, \dots, MSB_n)$\;

Initialize $S \gets 0$\;
\For{$i = 1$ \KwTo $n$}{
    \If{$(MSB_{max} - MSB_i) < T$}{
        $S \gets S + (a_i \times b_i)$ \tcp*{Include significant product}
    }
    \Else{
        \tcp*{Skip multiplication: $P_i$ is negligible}
    }
}
\Return $S$
\end{algorithm}
In the case of floating points represented in IEEE 754 format,
\begin{equation}
    \text{Value} = (-1)^S \times M \times 2^{E - \text{Bias}}
\end{equation}
The above algorithm can be used with a slight modification of using $E$ in place of $MSB$.
 The number of multiplications to be performed and the error is dependent on the distribution of data. Worst case absolute numerical error resulting from a singular conv\_approx() operation would be,
\begin{equation}
    (n - 1) \cdot \frac{1}{2^T} \cdot \max(a_i \times b_i)
\end{equation}
however, as seen in later sections, average error materialized using practical data sets is much smaller than that. When conv\_approx() is being used for inference using deep architectures with many layers, there is the possibility of large cumulative errors, however, the possibility is mostly mitigated as at each stage, both insignificant positive and negative products are discarded preventing errors from accumulating in either signed direction. As an extension, the threshold selection can be made adaptive and learning based, as the outcome is clearly dependent on the data distribution. It could also prove beneficial if threshold selection be made dynamic by setting a different threshold for each layer based on its weight distribution.          
This work explores using conv\_approx() in place of standard matrix convolution during CNNs' inference workloads. Since overfitting is usually caused by small weights which have picked up dataset specific characteristics, using conv\_approx() may mitigate the effects of overfitting by ignoring small products, however this is not validated.

\section{Hardware Implementation}
This section describes how a hardware implementation of the above algorithm was done to perform conv\_approx() operation specifically for input matrices of dimensions 4*4 and filter matrices of size 3*3 to test hardware feasibility. This operation is then iteratively called using inline assembly commands in C code to perform convolution for larger input matrices. To test effect of the approximation on the output of CNNs and as preliminary error analysis, conv\_approx() function was used to perform convolutions during inference of an MNIST image using LeNet-Accel model as described by S. Wang et al[13]. We choose LeNet-Accel because it uses only 3 * 3 and 4 * 4 convolution kernels. This standardized approach simplifies implementation and testing compared to the original LeNet, which uses varying matrix dimensions. More thorough error, accuracy, and performance analysis is described in subsequent sections. 
\subsection{Baseline Processor and Integration Strategy}
RISC-V is an open, modular instruction-set architecture built around a simple RISC-style base ISA, with a clean and extensible structure that separates a small mandatory core from optional standard extensions and reserved opcode spaces for customization, allowing designers to add application-specific instructions while remaining compatible with existing tools and software. The RI5CY core is a high-performance 32-bit RISC-V processor developed for the PULPino SoC within the Parallel Ultra-Low Power (PULP) platform, featuring a 4-stage in-order pipeline and full support for the RV32IMFC ISA, including integer multiplication, compressed instructions, and single-precision floating-point operations. In this work, one of the four currently unused opcodes, 0x77, is repurposed to define a custom conv\_approx() instruction, which is implemented through a custom hardware block integrated into the processor’s execution stage. This block interfaces directly with the instruction decoder to receive control signals and with the register file to access source operands, allowing the decoder to activate the specialized logic when the custom opcode is detected and to forward results through the standard write-back path, while its connection to the pipeline flow-control signals ensures correct synchronization with the core’s normal execution and memory access cycles.
\subsection{Custom Instruction Encoding and Execution Model}
The function conv\_approx() uses a inline RISC-V assembly to invoke a custom hardware instruction. The function is a R-type instruction encoded with opcode 0x77 via the .insn directive, the input register rs1 is used to store the size of the input array and the input register rs2 is used to store the starting address of the input array. The custom instruction consumes the two input registers holding address and size, performs some accelerator-defined operation which is elaborated in the next section, and writes its result into the output register result, which is mapped back to a C variable. The \_\_volatile\_\_ qualifier prevents the compiler from reordering or removing the instruction, ensuring that the custom operation is executed exactly as written.

\subsection{Hardware Architecture of the Approximate Convolution Unit}

A 5-stage Finite State Machine (FSM) in the accelerator module implements the \texttt{conv\_approx()} operation between an input $4 \times 4$ matrix and a $3 \times 3$ kernel matrix resulting in a $2 \times 2$ output matrix . Another custom instruction stores the 9 elements of the kernel matrix in the acceleration module, and each of their MSB locations are also computed and stored. The states of the FSM are \texttt{IDLE}, \texttt{GET\_DATA}, \texttt{STAGE\_1}, \texttt{STAGE\_2}, \texttt{STAGE\_3}, and \texttt{DONE}.

\noindent\textbf{\texttt{IDLE}:}
The FSM starts here, signaling that it is ready to accept a new command. It monitors
the input for a convolution operation signal, at which point it captures the
execution parameters before transitioning to data acquisition.

\vspace{0.5em}
\noindent\textbf{\texttt{GET\_DATA}:}
In this state, the FSM activates data fetching to fill its internal buffers.
It stays in this loop until the required amount of data is retrieved from
memory, ensuring that the \(4 \times 4\) data window is fully populated before
starting calculations.

\vspace{0.5em}
\noindent\textbf{\texttt{STAGE\_1} (\emph{MSB Analysis}):}
This stage computes the MSB locations of the 16 input values \((x_{i,j})\).
Negative values are converted to two's complement and processed by a
priority-encoder-like structure that extracts the MSB position of each
32-bit value as a 5-bit quantity. 

\vspace{0.5em}
\noindent\textbf{\texttt{STAGE\_2} (\emph{Pruning and Multiplication}):}
This stage performs the core optimization. For each output value,
\[
MSB_{\max} = \max \bigl(MSB(x_i) + MSB(w_i)\bigr)
\]
is computed using a combinational reduction tree. A partial product
\[
P_i = x_i \cdot w_i
\]
is evaluated only if
\[
MSB(x_i) + MSB(w_i) + T \ge MSB_{\max},
\]
where \(T\) is a tunable threshold; otherwise, the product is suppressed.

\vspace{0.5em}
\noindent\textbf{\texttt{STAGE\_3} (\emph{Accumulation}):}
The FSM sums the retained partial products for each convolution window to
produce the final outputs \((y_0, y_1, y_2, y_3)\).

\vspace{0.5em}
\noindent\textbf{\texttt{DONE}:}
The FSM asserts completion flags and waits for external acknowledgment before
resetting internal state and returning to \texttt{IDLE}.

{\small
\renewcommand{\arraystretch}{1.25}
\captionof{table}{Area, Power and Delay analysis using 65nm tech}
\begin{tabular}{|p{2.2cm}|p{2.8cm}|p{2.8cm}|}
\hline
\textbf{Metric} &
\textbf{MSB-based Pruning Block} &
\textbf{\% Increase w.r.t. RI5CY} \\
\hline
Area ($\mu$m$^2$) & 180050.68 & 108.23\% \\
\hline
Power (mW) & 0.849 & 11.50\% \\
\hline
Delay (ns) & 2.81 & 0\% \\
\hline
\end{tabular}
}

\subsection{Preliminary Error Analysis on inference using LeNet-Accel}

The following is a numerical comparison of one particular output of LeNet-Accel when done using standard convolution and using conv\_approx(). The tunable threshold has been set to approximate products that are less than 1\% of the largest product. The mean absolute error percentage is 0.97\% and the median absolute error percentage is 0.65\%. Errors are less than 1\% except for two low magnitude outliers, making them more susceptible to percentage changes. It is seen that it is extremely unlikely for the inference to change due to this approximation being performed at this threshold. More thorough analysis is performed in the subsequent section. 

{
\begin{center}
\centering
\captionof{table}{Comparison of Exact and Custom Block Outputs}
\label{tab:error_analysis}
\begin{tabular}{r r r l}
\hline
\textbf{Exact} & \textbf{Custom} & \textbf{Abs.} & \textbf{Percentage} \\
\textbf{Output} & \textbf{Output} & \textbf{Error} & \textbf{Error} \\
\hline
-3077  & -3073  & 4  & 0.13\% \\
-2351  & -2357  & 6  & 0.26\% \\
-8059  & -8122  & 63 & 0.78\% \\
-2502  & -2524  & 22 & 0.88\% \\
-5111  & -5137  & 26 & 0.51\% \\
6537   & 6632   & 95 & 1.45\% \\
-2352  & -2360  & 8  & 0.34\% \\
8202   & 8270   & 68 & 0.83\% \\
-1345  & -1400  & 55 & 4.09\% \\
16139  & 16206  & 67 & 0.41\% \\
\hline
\end{tabular}
\medskip
\end{center}
}

\section{Error Analysis and Results}
The conv\_approx() operation is first performed on individual MNIST images using various filters to examine output, error, and performance across varying approximation thresholds. Subsequently, inference is conducted on a LeNet-5 model trained on the MNIST dataset, where all convolution operations are replaced by conv\_approx(). This allows for a comprehensive analysis of the trade-offs between accuracy, computational cost (MACs), and error thresholds.

\subsection{MNIST}

The MNIST dataset (Modified National Institute of Standards and Technology)
is a massive collection of 70,000 small, grayscale images of handwritten
digits from 0 through 9. It is widely considered the ``Hello World'' of
machine learning because it provides a standardized way to train and test
computer vision models. Each image is $28 \times 28$ pixels, and the
data set is split into a training set of 60,000 images and a test set of
10,000 images. Since the optimization achieved by sparsity-dependent
paradigms such as \texttt{conv\_approx()} depends on the sparsity of the
input data set, the following table summarizes key statistics of MNIST.

{
\begin{center}
\captionof{table}{Statistics of MNIST dataset}
\label{tab:error_analysis}
\begin{tabular}{l l}
\hline
\textbf{Statistic} & \textbf{Value} \\
\hline
Pixels per image                     & 784 \\
Average zero pixels per image        & 633.92 \\
Minimum zero pixels per image        & 433 \\
Maximum zero pixels per image        & 750 \\
Standard deviation (zero pixels)     & 41.46 \\
Overall zero-pixel percentage        & 80.86\% \\
Mean               & $\approx 33.3$ \\
Standard deviation & $\approx 78.6$ \\
\hline
\end{tabular}
\end{center}
}

The following image demonstrates the visualization of the outputs produced by exact convolution and by \texttt{conv\_approx()} under different
approximation thresholds. Even for relatively large thresholds, the visual
features of the output remain largely indistinguishable. This is achieved
while realizing a 62.76\% reduction in the number of multiplications relative
to hard zero-skipping approaches and a 96.44\% reduction relative to exact
convolution.

{
\centering
\includegraphics[width=\columnwidth]{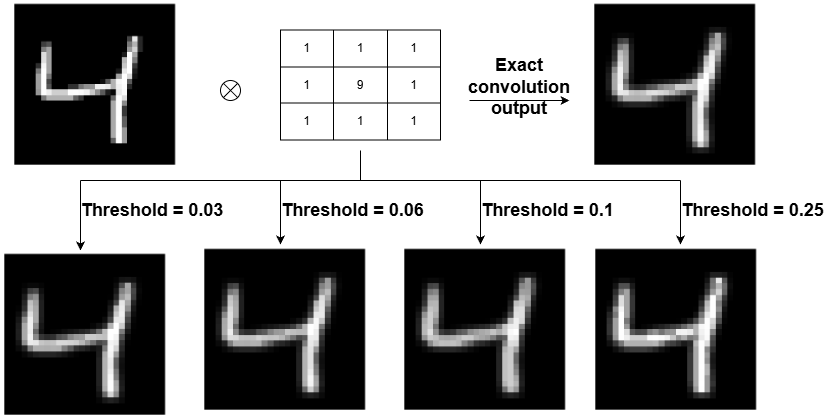}
\captionof{figure}{Visual demonstration of outputs with different error thresholds.}
\label{fig:example}
}
While the conv\_approx() operator may not be suitable for applications requiring an exact output, CNNs only require distinguishing between spatial patterns. Also, subsequent feature maps produce numerous small positive values which existing paradigms are forced to multiply and which conv\_approx() can skip. This makes conv\_approx() highly suitable during CNNs inference. The results of conv\_approx() during LeNet-5 inference are explored in the next section.

\begin{table}[H]
\centering
\caption{Average Number of Multiplications per Image}
\label{tab:mult_reduction_row}
\begin{tabular}{l r}
\hline
\textbf{Configuration} & \textbf{Avg. Multiplications} \\
\hline
Exact Convolution                  & 6084 \\
Non-Zero Multiplications           & 1354.0214 \\
Threshold = 0.03                   & 1256.661 \\
Threshold = 0.06                   & 1167.6030 \\
Threshold = 0.10                   & 1032.2543 \\
Threshold = 0.25                   & 506.2085 \\
\hline
\end{tabular}
\end{table}

{
\centering
\includegraphics[width=\columnwidth]{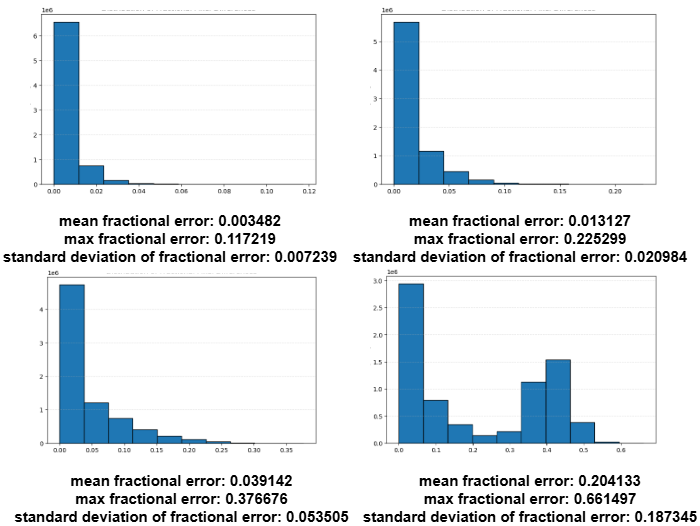}
\captionof{figure}{Distribution of fractional pixel errors for each of the four error thresholds. top left(T = 0.03),top right(T = 0.06), bottom left(T = 0.1), bottom right(T = 0.25).}
\label{fig:example}
}

\begin{table*}[t]
\centering
\caption{Impact of Different Thresholds on Convolution Multiplications and Accuracy for tanh and ReLU Activations}
\label{tab:threshold_analysis}
\begin{tabular}{llrrrrrrrr}
\toprule
& & \textbf{Exact} & \textbf{Non-Zero} & \multicolumn{6}{c}{\textbf{Threshold}} \\ \cmidrule{5-10}
& \textbf{Layer} & \textbf{Convolution} & \textbf{Multiplications} & \textbf{0.05} & \textbf{0.1} & \textbf{0.15} & \textbf{0.2} & \textbf{0.3} & \textbf{0.5} \\ \midrule

\multirow{4}{*}{\textbf{tanh}} & conv\_layer\_1 & 86400 & 20778.3 & 17964.25 & 16068.44 & 14829.27 & 12897.59 & 10537.48 & 7025.00 \\
& conv\_layer\_2 & 153600 & 153600.0 & 93101.69 & 74078.25 & 54322.51 & 45489.36 & 27699.23 & 12618.12 \\
& conv\_layer\_3 & 30720 & 30720.0 & 21629.76 & 16415.36 & 12688.72 & 9648.98 & 5552.83 & 1767.59 \\ \cmidrule{2-10}
& \textbf{Average MACs per Inference} & 270720 & 205098.3 & 132695.7 & 106562.05 & 81840.5 & 68035.93 & 43789.54 & 21410.71 \\
& \textbf{Accuracy} & $\sim$97--98\% & 97--98\% & 97.96\% & 97.53\% & 96.88\% & 97.62\% & 91.64\% & 61.38\% \\ \midrule

\multirow{4}{*}{\textbf{ReLU}} & conv\_layer\_1 & 86400 & 20778.3 & 17574.59 & 15637.01 & 14278.06 & 12234.10 & 10372.30 & 6988.47 \\
& conv\_layer\_2 & 153600 & 130788.0 & 66094.82 & 45949.18 & 42637.36 & 33567.69 & 17951.35 & 8410.60 \\
& conv\_layer\_3 & 30720 & 24688.8 & 13420.68 & 8691.39 & 5905.60 & 4146.67 & 3045.38 & 1116.28 \\ \cmidrule{2-10}
& \textbf{Average MACs per Inference} & 270720 & 176255.1 & 97090.09 & 70277.58 & 62821.02 & 49948.46 & 31369.03 & 16515.35 \\
& \textbf{Accuracy} & $\sim$97--98\% & $\sim$97--98\% & 97.55\% & 97.84\% & 97.86\% & 97.82\% & 97.39\% & 90.71\% \\ \bottomrule
\end{tabular}
\end{table*}

\subsection{LeNet-5 }
LeNet-5 is a classic convolution neural network composed of three convolution stages followed by two fully connected layers and a 10-class output. The first convolution layer (C1) applies 6 filters of size 5×5 to a single-channel input image, producing 6 feature maps that capture low-level patterns. The second convolution layer (C3) uses 16 filters of size 5×5 that operate on the 6 C1 feature maps to learn more complex, spatially distributed features. The third convolution layer (C5) applies 120 filters of size 4×4 over the 16 C3 maps, effectively collapsing the spatial dimensions and yielding a 120-dimensional feature vector. This representation is fed into a fully connected layer F6 with 84 neurons to perform high-level feature abstraction, followed by a final fully connected output layer that maps the 84 features to 10 class scores, typically corresponding to digit classes in MNIST. 
The sparsity in subsequent feature maps is dependent on the kind of activation function used. ReLU forces all negative values of the feature maps to 0, thus generating significant fractions of 0 values as analyzed in table. Other smooth activation functions, such as tanh generate no 0 values in subsequent feature maps, thereby restricting traditional sparsity exploiting paradigms to the ReLU activation function. Much of the non-zero values in these feature maps are small in magnitude, thereby making conv\_approx() suitable to skip significant number of multiplications.
\begin{table}[H] 
    \centering
    \footnotesize 
    \setlength{\tabcolsep}{1pt} 
    \renewcommand{\arraystretch}{0.9}
    \caption{Frequency of occurrence of absolute zero for ReLU LeNet-5 model}
    \label{tab:sparsity_lenet5}
    \begin{tabular}{lcccc}
        \toprule
        \textbf{Layer} & \textbf{Avg (\%)} & \textbf{Min (\%)} & \textbf{Max (\%)} & \textbf{StdDev (\%)} \\ 
        \midrule
        C1           & 45.11 & 38.43 & 54.11 & 2.52 \\
        S2           & 36.00 & 30.56 & 45.60 & 1.70 \\
        C3           & 50.25 & 44.73 & 56.25 & 1.58 \\
        S4           & 27.38 & 16.80 & 42.19 & 3.40 \\
        C5           & 55.10 & 41.67 & 65.00 & 3.45 \\
        F6           & 48.19 & 33.33 & 63.10 & 4.04 \\
        \bottomrule
    \end{tabular}
\end{table}

Accuracy-to-MAC-count trade-offs are characterized for LeNet using various thresholds, comparing the performance of tanh and ReLU activation layers. From the analysis, it is seen that using ReLU activation and at no loss of accuracy, inference is performed while having to compute on average only 11.58\% of the total multiplications, which translates to having to compute on average 17.79\% of the non-zero multiplications in convolutional layers. While using smoother activation functions like tanh, this translates to having to perform on average 25.13\% of total multiplications, which translates to 33.17\% of non-zero multiplications at no loss of inference accuracy.

{
\centering
\includegraphics[width=\columnwidth]{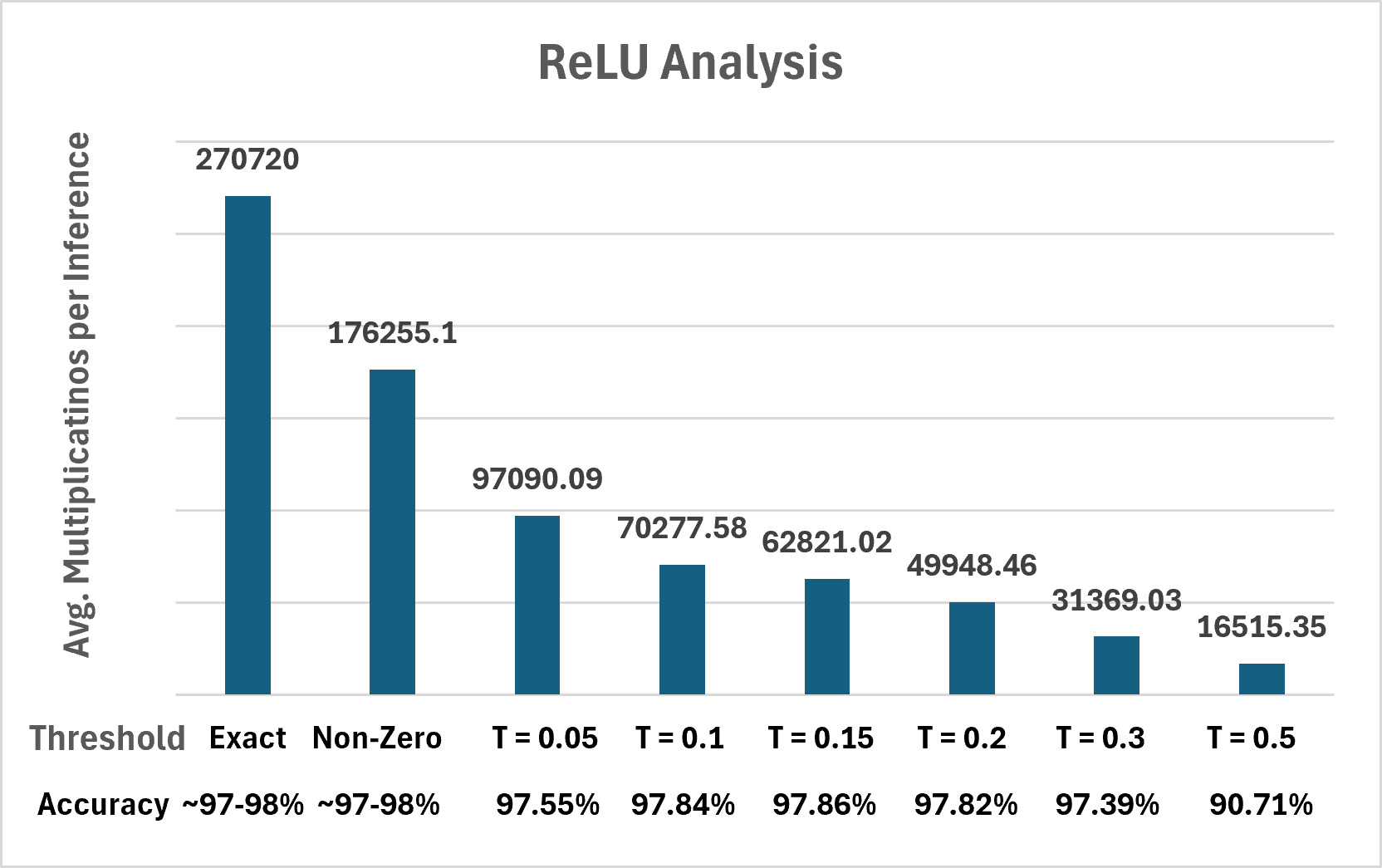}
\captionof{figure}{At T=0.3, 11.58\% of total MACs preserve accuracy.}
\label{fig:example}
}
{
\centering
\includegraphics[width=\columnwidth]{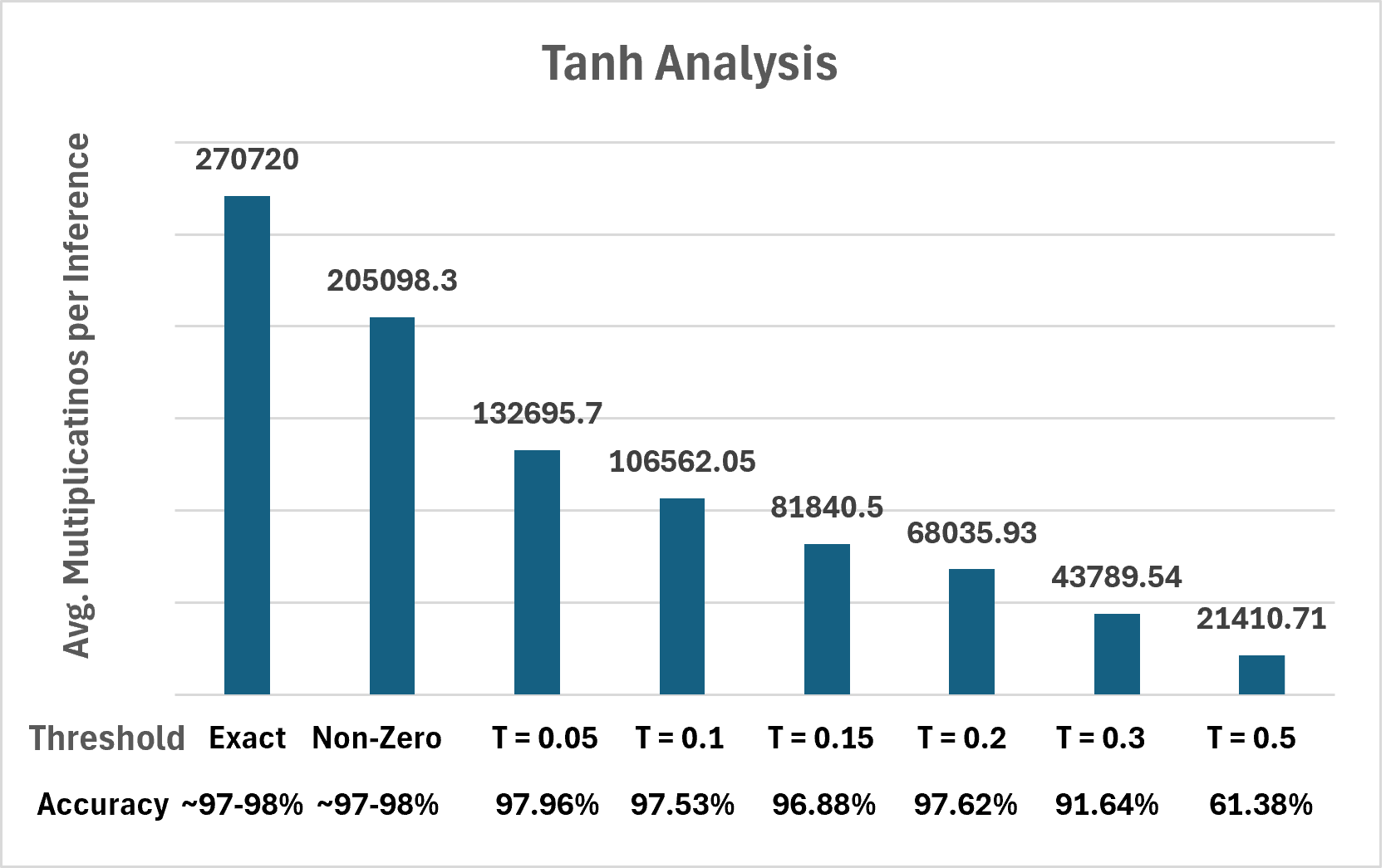}
\captionof{figure}{At T=0.2, 25.13\% of total MACs preserve accuracy.}
\label{fig:example}
}

\section{Conclusion}
This work has demonstrated that using this novel approximation algorithm, LeNet-5 inference can be performed without accuracy loss while achieving an 88.42\% and 74.87\% reduction in MAC operations for ReLU and tanh implementations, respectively. This in turn will lead to reduction in power consumed by clock gating of the multipliers when a product term need not be multiplied. As emphasized in the introduction, This power reduction will be sub-linear to the reduction in the number of MACs as in prior research, it has been established that in newer technologies memory access is the dominant consumer of power, which cannot be avoided. MAC operations have been quoted to consume a fraction of between 0.3 and 0.5 of the power consumption during inference according to various sources listed in the introduction. Assuming a conservative fraction of 0.4, the reduction in MAC operations will result in a reduction in power consumption of approximately 0.4 * 0.88 = 35.2\% and 0.4 * 0.74 = 29.96\% for ReLU and tanh activation, respectively. A possible way to realize greater power savings is to address the power consumed by memory access. Instead of fetching the 32 bit or 64 bit activation value and computing MSB location, 5 or 6 bit MSB locations can be pre-computed and stored in another 2-D array and fetched first, only fetch the value if the multiplication has to be performed. In CPUs, memory is often accessible by 32 bit or 64 bit chunks, however, specialized accelerators can be designed such that energy cost for 5 bit or 6 bit access be much lesser. In this way, the power savings are more linearized. The modified RTL code is available at \url{https://github.com/zeta-bot/Approximate_hardware_convolution}.

\end{document}